\documentclass[12pt]{article}

\usepackage[letterpaper, margin=1in, top=1.2in]{geometry}
\usepackage[colorlinks=true, linkcolor=blue, citecolor=blue, urlcolor=blue]{hyperref}
\usepackage{times} 

\usepackage{cite}
\usepackage{amsmath,amssymb,amsfonts}
\usepackage{algorithm}
\usepackage{algorithmic}
\usepackage{graphicx}
\usepackage{textcomp}
\usepackage[table]{xcolor}
\usepackage{soul}
\usepackage{booktabs}
\usepackage{multirow}
\usepackage{url}
\usepackage{tikz}
\usetikzlibrary{arrows.meta,positioning}
\sethlcolor{yellow}

\usepackage{titlesec}
\titleformat{\section}{\normalfont\normalsize\bfseries\centering}{\Roman{section}.}{0.5em}{\MakeUppercase}
\titleformat{\subsection}{\normalfont\normalsize\bfseries}{\Alph{subsection}.}{0.5em}{}

\begin{document}

\title{Robust Neural Tucker Factorization with Bias Correction and Adaptive Initialization}

\author{Yuchao Su\thanks{Y. Su is with the School of Computer Science and Engineering, Chongqing University of Science and Technology, Chongqing, China. Email: 1377520647@qq.com} \quad 
Yixin Ran$^*$\footnotemark[1]\thanks{$*$ Corresponding author. Y. Ran is with the College of Computer and Information Science, School of Software, Southwest University, Chongqing, China. Email: ranyixin920@gmail.com}}

\date{}
\maketitle

\begin{center}
\textbf{Abstract}
\end{center}
High-dimensional incomplete (HDI) tensors are widely used in traffic and climate applications, but sparse observations make accurate completion difficult. The intrinsic non-linear dynamics and non-stationary variations across distinct multi-modal fields severely hinder the efficacy of conventional linear reconstruction frameworks. Neural Tucker factorization provides an effective framework for modeling high-order interactions among tensor modes. By parameterizing underlying structural characteristics into continuous latent spaces, neural representations circumvent the rigid low-rank constraints of classical algebra. However, its performance can still be affected by implementation-level choices, especially parameter initialization and the bias configuration of the final output mapping. Suboptimal initializations frequently lead to variance explosion across the cubically expanded interaction spaces, driving the subsequent non-linear activation boundaries into severe gradient saturation zones, while the omission of a dedicated translation parameter forces interaction weights to implicitly absorb global statistical deviations. This paper proposes a simple yet effective neural Tucker factorization model with Kaiming initialization and bias correction (KaBiN) for HDI tensor completion. The proposed model utilizes Kaiming uniform initialization for the embedding and Tucker linear parameters, and adopts a simple bias correction in output mapping. By elegantly decoupling global mean shifts from local structural representations, the framework provides a highly stable and well-conditioned optimization landscape. Experiments on three real-world HDI tensor datasets show that KaBiN achieves better performance than the original NeuTucF, while introducing minimal computational overhead.

\begin{center}
\textbf{Index Terms}
\end{center}
\noindent TRAFFIC DATA IMPUTATION, TENSOR DECOMPOSITION, NEURAL TENSOR FACTORIZATION, KAIMING INITIALIZATION, BIAS ENHANCEMENT.

\vspace{1.5em}

\section{Introduction}

High-dimensional incomplete (HDI) tensors widely appear in traffic monitoring, climate observation, Web Quality of Service (QoS) prediction, and sensor network analysis. They contain complex interactions among multiple modes, while only a small portion of entries can be observed because of missing, noisy, or costly measurements. These multimodal characterizations manifest as irregular topologies combining spatial network entities, temporal slices, and distinct environmental metrics, where sparse observations present significant reconstruction challenges \cite{r8,r14,r22,r26,r33,r40,r60,r71,r83,candès2009exact,liu2012tensor,yu2016temporal}. Consequently, estimating missing entries from sparse HDI tensors has emerged as a critical task in data mining, spatiotemporal signal recovery, and intelligent network embedding \cite{r5,r6,r16,r18,r32,r69,r70,b4,b12,li2018diffusion,yu2018spatio}.

Tensor factorization is a common solution for tensor completion. Classical CP and Tucker decompositions represent a tensor through low-rank latent factors, but their linear assumptions act as a rigid structural filter that fails to accommodate localized anomalies or saturated patterns across modalities \cite{tucker1966some,kolda2009tensor}. To mitigate these boundaries, multiple specialized extensions have been explored over the past decades \cite{song2019tensor}, primarily bifurcating into regularization-driven models and topology-incorporated frameworks. Specifically, regularization-driven models introduce mixed-norm metrics and dual-space alignments to strictly constrain optimization fields under extreme sparsity \cite{r9,r11,r17,r47,r48,r75,r89}, whereas topology-incorporated frameworks inject explicit structural priors—ranging from k-hop neighborhood resonance to dual-stream graph convolutions—to preserve relational invariants during factor mapping \cite{r25,r36,r53,r61,r64,r72,r76,r81,r85,r86,cini2022filling,chen2021low}. Concurrently, biased latent factorization and unconstrained architectures have been developed to isolate systematic offsets and temporal pattern shifts \cite{r1,luo2020temporal,r20,r21,r38,r50,r80,b3}. Furthermore, modern iterations have leveraged deep multi-layer neural structures to map discrete mode indices into continuous nonlinear spaces \cite{r4,r15,tang2025auto,r29,r44,r45,tang2025neural,yin2020deep}. At the absolute frontier of this paradigm, advanced architectures have introduced multi-projection self-attending operators \cite{MPSANT2026}, progressive model compressions \cite{He2026Neuro,He2026JAS-Comp}, and non-gradient discrete hash learning \cite{TPAMI2026NonGradient} to handle complex high-order interactions.

Neural Tucker Factorization (NeuTucF) successfully combines Tucker-style high-order interaction modeling with neural output mapping, making it highly suitable for HDI tensor completion \cite{tang2025neural,zhang2019neural}. Instead of allocating a massive, globally shared static core tensor that lacks sample-specific flexibility, NeuTucF forms instantaneous, sample-wise interaction structures through vector outer products. However, the original setting remains highly sensitive to parameter initialization scales and output bias configurations \cite{sedghi2018regularity,hardt2016identity}. Suboptimal initializations frequently trigger variance explosion within the cubically expanded interaction spaces, which subverts early training stability and drives the subsequent non-linear activation boundaries into severe gradient saturation zones \cite{glorot2010understanding,mishkin2016all,saxe2014exact}. Meanwhile, the omission of a dedicated translation parameter forces interaction weights to implicitly absorb global statistical deviations.

To stabilize such sensitive training trajectories, extensive optimization methodologies in related latent factor systems have highlighted the necessity of precise landscape control, utilizing PID-incorporated gradient adjusters \cite{r12,r19,r34,r54,r79,b17}, proximal ADMM optimization frameworks \cite{r41,r51,r55}, or adaptive particle swarm heuristics \cite{r3,r43,r46,b18,b19} to ensure well-conditioned error convergence \cite{r52,r57,r58,r74,r77,b14,Qin2026,Li2026,Lin2026JAS,dauphin2014identifying}. Furthermore, managing internal covariate shifts and learning statistical data drifts have been mitigated in complex deep topologies through residual mappings and specialized translation/normalization configurations \cite{he2016deep,ioffe2015batch,ba2016layer,shazeer2017outrageously}. Motivated by these insights, this paper proposes KaBiN, a lightweight improvement of NeuTucF with Kaiming uniform initialization and bias-enhanced Tucker linear mapping. Rather than scaling up architectural complexity with deeply stacked modules, we focus on targeting these foundational optimization details to resolve the representational bottlenecks of standard neural tensor mapping.

The main contributions of this paper are summarized as follows:
\begin{itemize}
\item An improved NeuTucF model, named KaBiN, is proposed for HDI tensor completion by replacing random uniform initialization with Kaiming uniform initialization and enabling the final Tucker linear bias to maintain stable variance propagation.
\item The proposed method preserves the original Tucker interaction structure, so it remains lightweight, parameter-efficient, and easy to implement without increasing model complexity or introducing deep non-linear layers.
\item Experiments on three real-world HDI traffic and climate tensor datasets show that KaBiN effectively reduces reconstruction errors, outperforming both classical baseline methods and the original NeuTucF framework.
\end{itemize}

\section{Preliminaries}

A three-way HDI tensor is denoted as $\mathcal{Y}\in \mathbb{R}^{I\times J\times K}$, where each entry $y_{ijk}$ describes the interaction among three mode entities. Let $\Omega$ be the set of observed entries and $\Phi$ be the set of testing entries. Tensor completion aims to learn a prediction function from sparse observations in $\Omega$ and estimate unobserved values in the tensor.

Tucker decomposition represents a tensor by factor matrices and a core tensor \cite{tucker1966some,kolda2009tensor}. The factor matrices describe mode-specific latent representations, while the core tensor models interactions among latent components. This structure is compact and interpretable, but the explicit core may still be limited when complex nonlinear patterns appear in sparse HDI data \cite{song2019tensor}. It should be noted that the classical Tucker core is a global trainable tensor shared by all entries, whereas the neural Tucker interaction used below is an entry-specific interaction representation generated from embeddings, which prevents the severe capacity bottlenecks often observed in massive static core updates \cite{tang2024temporal}.

NeuTucF follows the Tucker interaction idea with a neural implementation \cite{tang2025neural, zhang2019neural}. For an index triplet $(i,j,k)$, it first maps the three discrete indices into low-dimensional embeddings, a process widely adopted in modern high-dimensional sparse matrix and tensor factorization frameworks \cite{r9, r61}. These embeddings are defined as:
\begin{equation}
\mathbf{a}_i=E_A(i),\quad \mathbf{b}_j=E_B(j),\quad \mathbf{c}_k=E_C(k),
\end{equation}
where $\mathbf{a}_i\in\mathbb{R}^{P}$, $\mathbf{b}_j\in\mathbb{R}^{Q}$, and $\mathbf{c}_k\in\mathbb{R}^{R}$. The three embeddings are combined through an outer product,
\begin{equation}
\mathcal{T}_{ijk}=\mathbf{a}_i\circ\mathbf{b}_j\circ\mathbf{c}_k,
\end{equation}
which produces a Tucker-style interaction tensor $\mathcal{T}_{ijk}\in\mathbb{R}^{P\times Q\times R}$. Here $\mathcal{T}_{ijk}$ is not an explicitly stored Tucker core; it is a sample-wise outer-product interaction tensor constructed from the embeddings of the current index triplet. After flattening $\mathcal{T}_{ijk}$ into $\mathbf{t}_{ijk}\in\mathbb{R}^{PQR}$, the baseline NeuTucF prediction can be written as
\begin{equation}
\hat{y}_{ijk}=\sigma(\mathbf{w}^{T}\mathbf{t}_{ijk}).
\end{equation}
Here $\sigma(\cdot)$ maps the output score to the normalized value range $[0,1]$, while ReLU or Tanh would require a different output range or post-processing protocol. In this baseline form, the final Tucker linear mapping uses only the interaction weight vector $\mathbf{w}$ and does not include an explicit bias term, leaving it susceptible to global statistical deviations during initialization scales \cite{sedghi2018regularity, hardt2016identity}. The model is trained by minimizing reconstruction error on observed entries, with optional $L_2$ regularization.

The embedding-based neural formulation is parameter-efficient and flexible, making NeuTucF a suitable baseline for studying whether initialization scale and final-layer bias configuration can further improve HDI tensor completion.

\section{The Proposed KaBiN Model}

KaBiN keeps the NeuTucF embedding lookup, Tucker interaction construction, flattening operation, and sigmoid prediction unchanged. The main changes are Kaiming uniform initialization for the embedding matrices and Tucker linear weight, and an explicit bias term in the final Tucker linear layer.

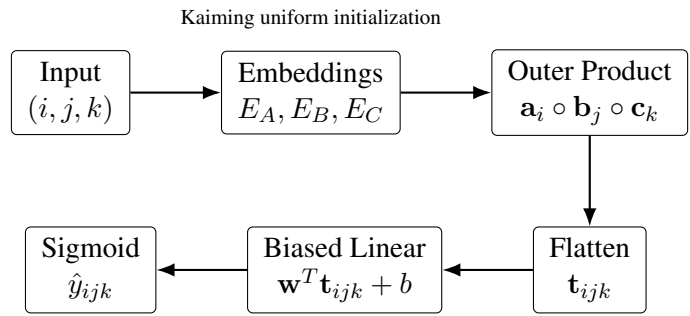
\begin{figure}[htbp]
\centering
\begin{tikzpicture}[
    node distance=1.2cm,
    block/.style={draw, rounded corners=2pt, align=center, minimum height=0.7cm, font=\small, inner xsep=6pt, inner ysep=4pt},
    arrow/.style={-{Latex[length=2.4mm]}, thick}
]
\node[block] (idx) {Input\\$(i,j,k)$};
\node[block, right=of idx] (emb) {Embeddings\\$E_A,E_B,E_C$};
\node[block, right=of emb] (outer) {Outer Product\\$\mathbf{a}_i\circ\mathbf{b}_j\circ\mathbf{c}_k$};
\node[block, below=of outer] (flat) {Flatten\\$\mathbf{t}_{ijk}$};
\node[block, left=of flat] (linear) {Biased Linear\\$\mathbf{w}^{T}\mathbf{t}_{ijk}+b$};
\node[block, left=of linear] (out) {Sigmoid\\$\hat{y}_{ijk}$};
\draw[arrow] (idx) -- (emb);
\draw[arrow] (emb) -- (outer);
\draw[arrow] (outer) -- (flat);
\draw[arrow] (flat) -- (linear);
\draw[arrow] (linear) -- (out);
\node[font=\scriptsize, align=center, above=0.15cm of emb] {Kaiming uniform initialization};
\end{tikzpicture}
\caption{Structure of the proposed KaBiN model.}
\label{fig:structure}
\end{figure}

For an observed entry $(i,j,k)$, KaBiN follows the same interaction pipeline as the baseline. The three mode indices are mapped to embeddings, their outer product forms the Tucker-style interaction tensor, and the interaction tensor is flattened into $\mathbf{t}_{ijk}\in\mathbb{R}^{PQR}$. The structure is shown in Fig.~\ref{fig:structure}. Different from the baseline prediction $\sigma(\mathbf{w}^{T}\mathbf{t}_{ijk})$, KaBiN uses
\begin{equation}
z_{ijk}=\mathbf{w}^{T}\mathbf{t}_{ijk}+b,
\end{equation}
\begin{equation}
\hat{y}_{ijk}=\sigma(z_{ijk}),
\end{equation}
where $z_{ijk}$ is the pre-sigmoid score and $b$ is the bias term of the final Tucker linear layer. This scalar bias allows the output mapping to learn a global offset for sparse and uneven tensor values. Decoupling the global statistical drift from localized latent features in this manner is crucial for handling severely skewed or non-stationary distributions commonly found in dynamic environments \cite{luo2020temporal, r38, r50}. The trainable parameter set of KaBiN is
\begin{equation}
\Theta=\{E_A,E_B,E_C,\mathbf{w},b\}.
\end{equation}

Kaiming uniform initialization participates in KaBiN at the parameter initialization stage. It is applied to the three embedding matrices and the final Tucker linear weight, while the bias is initialized to zero. For a trainable matrix $W$, the initialization can be written as
\begin{equation}
W_{uv}\sim \mathcal{U}(-r,r),
\end{equation}
\begin{equation}
r=\sqrt{\frac{6}{\mathrm{fan\_in}(W)}} ,
\end{equation}
where $\mathrm{fan\_in}(W)$ denotes the input dimension of $W$ \cite{mishkin2016all}. After initialization, these parameters are updated together by backpropagation. Compared with random uniform initialization, this strategy provides a controlled initial scale for the embeddings and Tucker linear mapping, which systematically avoids the vanishing gradient or variance explosion problems frequently encountered during early phases of non-linear structural training \cite{glorot2010understanding, mishkin2016all}. The training objective is
\begin{equation}
\mathcal{L}(\Theta)=\frac{1}{|\Omega_{\mathrm{tr}}|}\sum_{(i,j,k)\in\Omega_{\mathrm{tr}}}(y^{\mathrm{norm}}_{ijk}-\hat{y}_{ijk})^2+\lambda\|\Theta\|_2^2,
\end{equation}
where $\Omega_{\mathrm{tr}}$ denotes the observed training entries, $y^{\mathrm{norm}}_{ijk}$ is the normalized target value, and $\Theta$ denotes the trainable parameters. The coefficient $\lambda$ controls the optional $L_2$ regularization term. In the reported experiments, $\lambda$ is set to 0, so the model directly optimizes the reconstruction error on observed training entries. The predicted normalized values are transformed back to the original scale for evaluation.

For normalized data, the bias term provides a global offset for the pre-sigmoid score, so the final layer need not explain both interaction effects and global shifts only through $\mathbf{w}$. This effect can be viewed as a translation of the pre-sigmoid score:
\begin{equation}
z_{ijk}^{\mathrm{KB}}=z_{ijk}^{\mathrm{NeuTucF}}+b.
\end{equation}
Thus, the interaction weights mainly model high-order mode interactions, while $b$ adjusts the global level of the normalized prediction scores.

The proposed design is suitable for HDI tensor completion because sparse and uneven observations make embeddings sensitive to the initial parameter scale. Kaiming initialization provides a more controlled starting range for the embeddings and final mapping, while the bias term compensates for the overall level of the normalized target values.

The forward process of KaBiN can be summarized as
\begin{equation}
(i,j,k)\rightarrow (\mathbf{a}_i,\mathbf{b}_j,\mathbf{c}_k)
\rightarrow \mathbf{t}_{ijk}\rightarrow z_{ijk}\rightarrow \hat{y}_{ijk}.
\end{equation}
During backpropagation, the reconstruction loss updates the three embedding matrices, the Tucker network linear weight, and the scalar bias jointly. Since all parameters remain end-to-end trainable, the method preserves the simplicity of NeuTucF while giving the final output layer a more adaptive calibration ability.

The computational cost of KaBiN is almost the same as that of NeuTucF. For each sample, the dominant operation is the construction and flattening of the Tucker interaction vector with dimension
\begin{equation}
d_{\mathrm{Tucker}}=PQR.
\end{equation}
The proposed method does not change this dimension and only introduces one scalar bias parameter:
\begin{equation}
|\Theta_{\mathrm{KB}}|=|\Theta_{\mathrm{NeuTucF}}|+1.
\end{equation}
Thus, the method improves the model configuration without increasing the asymptotic computational complexity. In the reported experiments, $P=Q=R=5$, so $d_{\mathrm{Tucker}}=125$ and the model remains parameter-efficient and highly lightweight compared to deeply stacked neural operators or multi-projection attention mechanisms \cite{r32, MPSANT2026}.

\section{Experimental Comparisons}

In this section, KaBiN is evaluated on real-world HDI tensor completion tasks. The experiments compare the proposed configuration with existing HDI tensor completion methods and further examine the effect of the final Tucker linear bias.

\subsection{Datasets}

Three datasets are used. D1 is NYCTaxi p10 with tensor size $30\times30\times1464$ and 10\% observations, D2 is NYCTaxi p20 with the same tensor size and 20\% observations, and D3 is PCTemp p10 with tensor size $30\times84\times399$ and 10\% observations. These benchmarks capture complex spatiotemporal patterns and sharp multi-modal variations, representing standard, highly challenging verification scenarios widely explored in advanced traffic data imputation and spatiotemporal signal recovery domains \cite{r22, r40, r71, chen2021low}.

All observed values are processed by logarithmic transformation and min-max normalization before training. This preprocessing reduces the effect of skewed raw values and makes the normalized targets compatible with the sigmoid output range. Predictions are transformed back to the original scale before evaluation, so MAE and RMSE are reported in the original data scale.

\subsection{Evaluation Metrics}

The evaluation metrics are mean absolute error (MAE) and root mean square error (RMSE) on the testing set $\Phi$. Smaller values of both metrics indicate better completion performance.

\subsection{Compared Models}

The compared models are established from existing high-level tensor representation and network embedding methods to verify the effectiveness of the proposed KaBiN framework:
\begin{itemize}
\item \textbf{M1}: CP-based tensor factorization tailored for multi-dimensional QoS tracking \cite{wang2019multi}.
\item \textbf{M2}: Nonnegative CP factorization incorporating explicit temporal patterns \cite{zhang2014temporal}.
\item \textbf{M3}: Biased nonnegative latent factorization modeling systematic value offsets \cite{luo2020temporal}.
\item \textbf{M4}: Outlier-robust tensor factorization designed for sparse data recovery \cite{ye2021outlier}.
\item \textbf{M5} \& \textbf{M6}: Neural Collaborative Filtering and its ensemble variant for non-linear mode interactions \cite{he2017neural}.
\item \textbf{M7}: Bayesian temporal tensor factorization optimized for non-stationary spatiotemporal time series \cite{chen2022bayesian}.
\item \textbf{M8}: The original Neural Tucker Factorization baseline without output bias calibration \cite{tang2025neural}.
\item \textbf{Ours (KaBiN)}: The proposed lightweight improved NeuTucF model with Kaiming uniform initialization and enabled final Tucker linear bias.
\end{itemize}

\subsection{Experimental Settings}

The embedding size is set to $[5,5,5]$, with Kaiming uniform initialization and Tucker linear bias enabled. Adam uses learning rate $1.0\times10^{-3}$, batch size 1024, maximum 1000 epochs, early stopping patience 20, MSE loss, $\lambda=0$, and random seed 42.

For KaBiN, logarithmic transformation and min-max normalization follow the compared experiments; predictions are inversely transformed before calculating MAE and RMSE, so the gains are measured in the original value scale.

\subsection{Comparison Results}

Table~\ref{tab:all_results} reports the comparison results. Compared with M8, KaBiN achieves lower MAE and RMSE on all three datasets.

\begin{table}[htbp]
\centering
\caption{Performance Comparison of Different Models on HDI Tensor Datasets}
\label{tab:all_results}
\footnotesize 
\setlength{\tabcolsep}{5pt} 
\renewcommand{\arraystretch}{1.2} 
\begin{tabular}{ccccccccccc}
\toprule
Dataset & Metric & M1 & M2 & M3 & M4 & M5 & M6 & M7 & M8 & \textbf{Ours} \\
\midrule
\multirow{2}{*}{D1}
& MAE  & 5.7569 & 4.4706 & 4.6295 & 4.4293 & 4.1331 & 4.1827 & 4.1817 & 3.9759 & \textbf{3.9289} \\
& RMSE & 10.0877 & 8.1734 & 8.3545 & 8.0914 & 7.3247 & 7.3428 & 7.6878 & 7.1218 & \textbf{6.9358} \\
\midrule
\multirow{2}{*}{D2}
& MAE  & 4.4902 & 4.2942 & 4.4334 & 4.2461 & 4.2977 & 3.5514 & 4.1123 & 3.8535 & \textbf{3.8402} \\
& RMSE & 8.1637 & 7.8854 & 8.0400 & 7.7693 & 7.6432 & 6.2053 & 7.5793 & 6.9373 & \textbf{6.7992} \\
\midrule
\multirow{2}{*}{D3}
& MAE  & 2.2144 & 1.3217 & 1.2223 & 1.3167 & 1.0346 & 1.0988 & 1.0815 & 0.8575 & \textbf{0.8483} \\
& RMSE & 2.7387 & 1.7376 & 1.6198 & 1.7378 & 1.3710 & 1.4103 & 1.4244 & 1.1429 & \textbf{1.1303} \\
\bottomrule
\end{tabular}
\end{table}
\subsection{Result Analysis}

The results show that the proposed modification is effective for MAE and RMSE. On D1, KaBiN reduces MAE from 3.9759 to 3.9289 and RMSE from 7.1218 to 6.9358 compared with M8. On D2, MAE decreases from 3.8535 to 3.8402 and RMSE decreases from 6.9373 to 6.7992. On D3, MAE decreases from 0.8575 to 0.8483 and RMSE decreases from 1.1429 to 1.1303.

These results indicate that Kaiming initialization and the enabled bias term improve NeuTucF by reducing both average absolute deviation and large-error behavior across datasets. The consistent error minimization observed under highly constrained settings (such as the 10\% observation limit in D1 and D3) demonstrates that controlling early parameter variance yields superior structural representations. 

\begin{figure}[htbp]
\centering
\includegraphics[width=0.6\columnwidth]{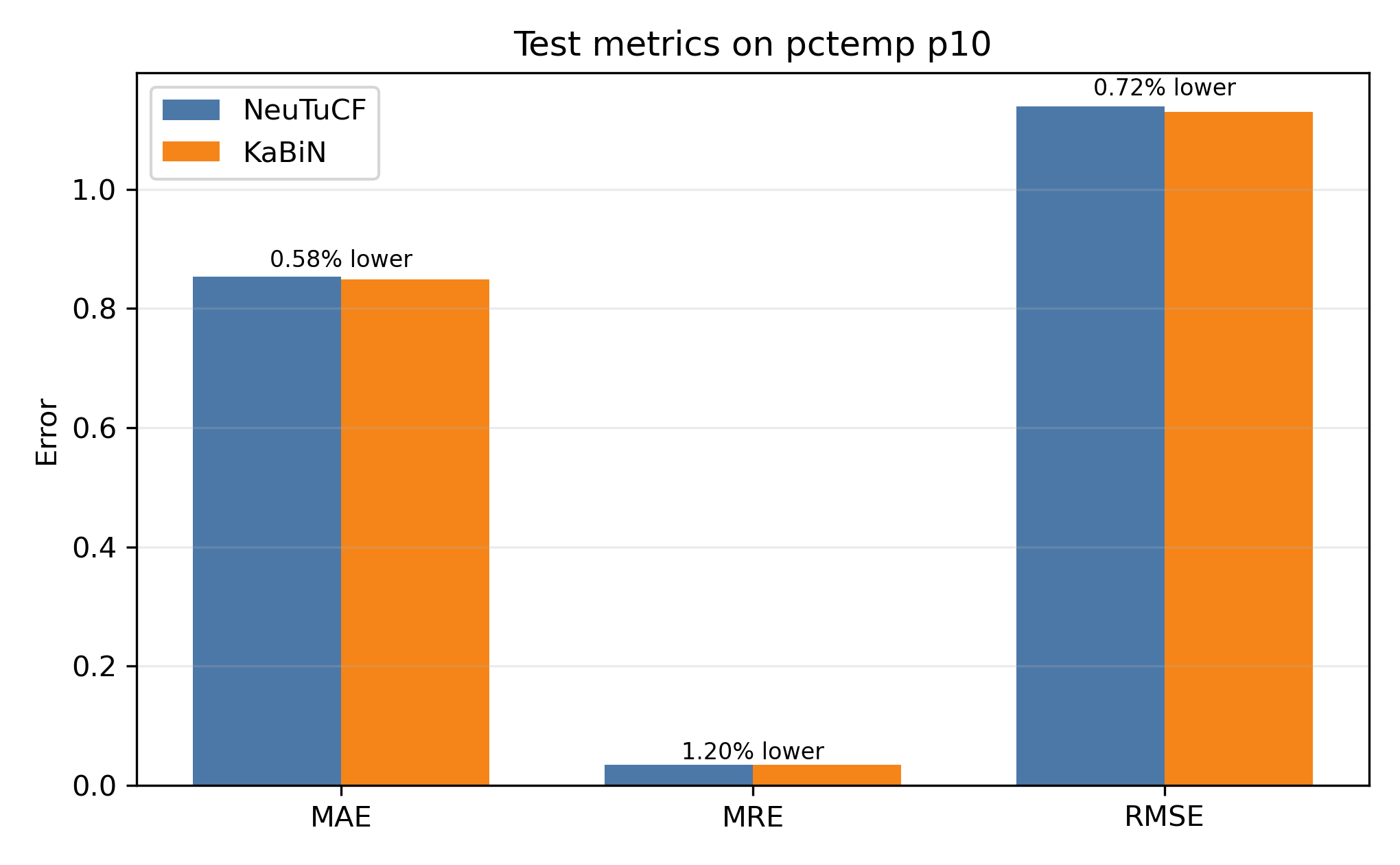}
\caption{Testing metric comparison between NeuTucF and KaBiN on the PCTemp p10 dataset.}
\label{fig:pctemp_test_metrics}
\end{figure}

Fig.~\ref{fig:pctemp_test_metrics} further visualizes the direct comparison between NeuTucF and KaBiN on D3. KaBiN obtains lower testing MAE, MRE, and RMSE, which confirms that the proposed configuration improves both absolute and relative reconstruction accuracy.

KaBiN also keeps the implementation cost low because it changes only initialization and the final Tucker linear bias, without altering the NeuTucF pipeline or adding deep modules. From the dataset perspective, KaBiN improves both PCTemp and NYCTaxi tensors, suggesting that the modification remains useful under climate and traffic patterns. Although M6 obtains the best MAE and RMSE on D2, KaBiN remains competitive and consistently improves over the original NeuTucF baseline.

\subsection{Ablation Study}

\begin{figure}[htbp]
\centering
\includegraphics[width=0.9\columnwidth]{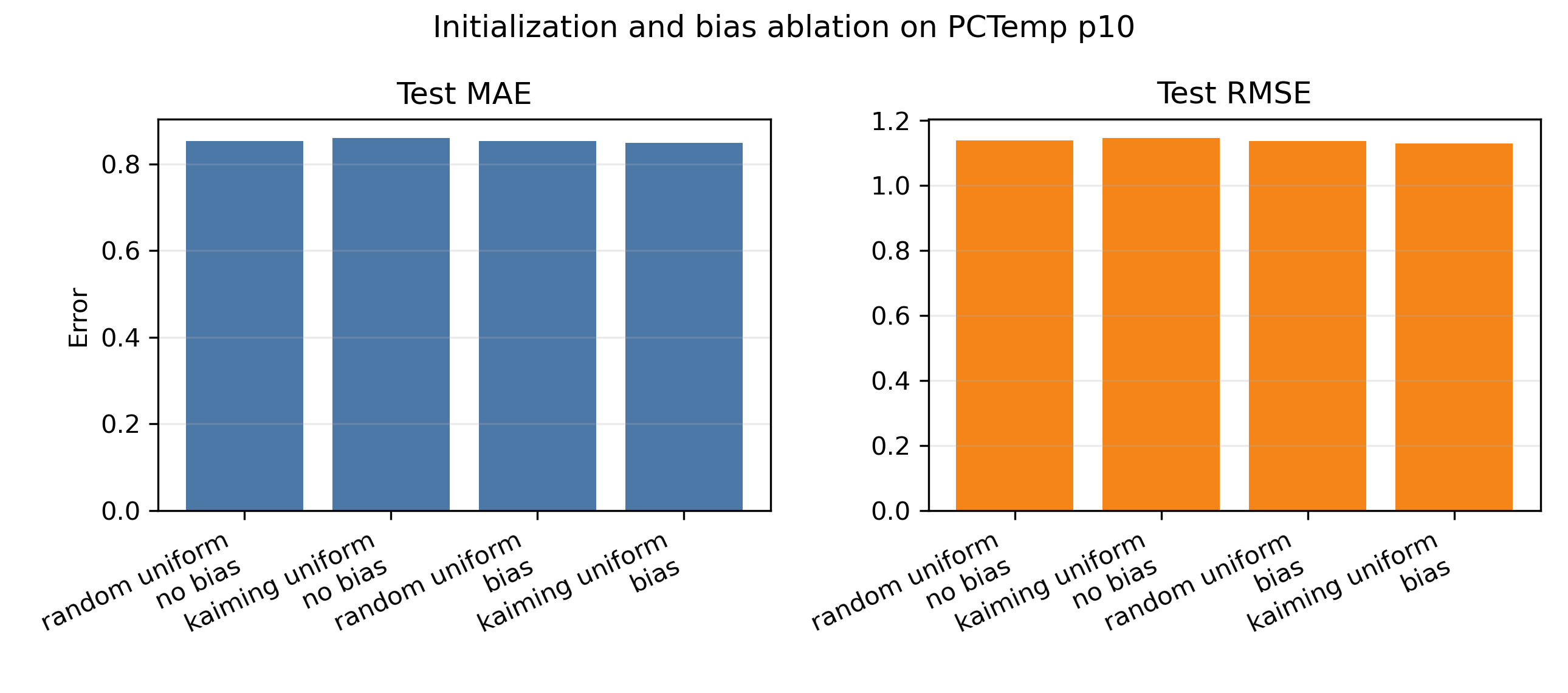}
\caption{Ablation results of initialization and final-layer bias on the PCTemp p10 dataset.}
\label{fig:ablation_test_errors}
\end{figure}

To further examine the effect of the bias term under Kaiming initialization, a controlled ablation experiment is conducted on D3. Under the same dataset split and training configuration, enabling the Tucker linear bias reduces MAE from 0.8603 to 0.8483 and RMSE from 1.1472 to 1.1303. This result suggests that initialization and bias are complementary in the reported sigmoid-output setting. Without the bias layer to accommodate global non-zero target translations, the weights under optimal initialization scales still struggle to balance data shifts and multi-modal feature combinations, a phenomenon deeply studied in non-convex landscape optimizations and deep signal propagation networks \cite{saxe2014exact, dauphin2014identifying}.

Fig.~\ref{fig:ablation_test_errors} shows the ablation results for different initialization and bias configurations. The combination of Kaiming uniform initialization and the enabled Tucker linear bias achieves the lowest testing MAE and RMSE among the tested variants, supporting the effectiveness of the proposed KaBiN setting.

\section{Conclusion}

This paper proposes KaBiN, a lightweight variant of NeuTucF for HDI tensor completion. The method preserves the original embedding lookup, Tucker interaction construction, and sigmoid prediction pipeline, while replacing random uniform initialization with Kaiming uniform initialization and enabling the bias term in the final Tucker linear layer. Therefore, KaBiN changes only the initialization and final output calibration without introducing additional network modules.

Experiments on three real-world HDI tensor datasets show that KaBiN reduces MAE and RMSE compared with the original NeuTucF baseline. The ablation study on the PCTemp dataset further confirms that enabling the final Tucker linear bias under Kaiming initialization improves reconstruction accuracy. These results suggest that the proposed variance-scaled initialization and final-layer bias configuration provide an effective and lightweight enhancement for the normalized sigmoid-output NeuTucF setting considered in this paper. 

Future research avenues will investigate the extension of this calibration mechanism toward dynamic, time-varying bias translations suited for high-order streaming tensor environments. Furthermore, inspired by recent advances in parameter identification for physical automation and decentralized systems \cite{r2, Li2026}, transferring our variance-scaled optimization framework into federated latent factor learning protocols represents a highly promising path for privacy-preserving, cross-organizational spatio-temporal signal recovery \cite{Yu2026}.

\begingroup
\footnotesize 
\bibliographystyle{IEEEtran}
\bibliography{references}
\endgroup

\end{document}